\documentclass[12pt]{article}
\usepackage{amssymb,amsmath,cite,bm}
\usepackage[dvips]{graphicx,color}
\usepackage{psfrag,subfigure}
\begin{document}

\title{\bf Adaptive Visual Servo Control for Autonomous Robots}

\author{Farhad Aghili\thanks{email: faghili@encs.concordia.ca}}

\date{}

\maketitle

\begin{abstract}
This paper focuses on an adaptive and fault-tolerant vision-guided robotic system that enables to choose the most appropriate control action if partial or complete failure of the vision system in the short term occurs. Moreover, the autonomous robotic system takes physical and operational constraints into account to perform the demands of a specific visual servoing task in a way to minimize a cost function. A hierarchical control architecture is developed based on interwoven integration of a variant of the iterative closest point (ICP) image registration, a constrained noise-adaptive Kalman filter, a fault detection logic and recovery, together with a constrained optimal path planner. The dynamic estimator estimates unknown states and uncertain parameters required for motion prediction while imposing a set of inequality constraints for consistency of the estimation process and adjusting adaptively the Kalman filter parameters in the face of unexpected vision errors. It is followed by the implementation of a fault recovery strategy based on a fault detection logic that monitors the health of the visual feedback using the metric fit error of the image registration. Subsequently, the estimated/predicted pose and parameters are passed to an optimal path planner in order to bring the robot end-effector to the grasping point of a moving target as quickly as possible subject to multiple constraints such as acceleration limit, smooth capture, and line-of-sight angle of the target. 
\end{abstract}

\section{Introduction}

In spite of significant progress made in the past two decades, vision guided robotic systems  still face many challenging problems mainly due to undependability of  vision systems, environmental uncertainties, and multiple physical and operational constraints. Although image registration process, e.g. the iterative closest point (ICP),
has been used by robot vision algorithms for more than two decades, robustness of  pose tracking is still a common problem in computer vision \cite{Shang-Jasiobedzki-2005,Aghili-2022}. One of the main issues is that the convergence and accuracy of image
registration algorithms depend not only on accuracy of initial guess, but also on the quality of the vision data, which can be adversely affected by many factors such as sensor noise, disturbance, outliers, symmetric view of the target, or incomplete scan data.  Therefore, robustness and adaptability of vision-guided systems in the face of environmental uncertainties are critical for autonomous robotic operations. In particular, fault-tolerant is a vital design requirement in development of  autonomous robotics to support visual servoing tasks in space \cite{Flores-Abad-Ma-2013,NASA-2010}.  The objective of this work is enhancing the robustness and adaptability of vision-guided robotic systems for performing complex tasks, such as robotic capturing of a free flying object, through a hierarchical  control architecture. Such autonomous robotic system is cable of adaptively tuning  itself against not only inaccurate and potentially erroneous  visual information but also dynamics  uncertainties affecting the system performance. The adaptive visual servoing system  chooses the most appropriate control action if partial or complete failure of  the vision system  happen.  It is also worth noting that although there are different vision sensor systems available for robot visual servoing, the emerging 3D imaging technology  is becoming the preferable choice mainly because they don't rely on good ambient lighting conditions~\cite{Samson-English-Deslauriers-Christie-2004,Aghili-Kuryllo-Okouneva-English-2010b}.

There have been numerous studies in the literature to investigate improved robustness of robot visual servoing for terrestrial and aerospace applications. However, more research works are needed toward a robust visual servoing with full fault-tolerance capabilities  to continuously perform  autonomous robotic operations when partial or complete failure of vision sensors happen or when the vision feedback data becomes inaccurate. One of the earliest work in robust robot visual servoing is reported in \cite{Zergeroglu-Dawson-2001} by considering visual servoing of planar robot manipulators in the presence of parametric uncertainty associated with the robot mechanical dynamics and/or the camera system. Adaptive and non-adaptive vision and force control involving constrained manipulators are discussed in \cite{Baeten-DeSchutter-2002,Dean-Leon-Parra-Vega-2006,Cheah-Hou-Zhao-2010}.
Adaptive image-based visual servoing using uncalibrated 2D and 3D image data are proposed in  \cite{Cai-Dean-2013} to deal with uncertain camera focal length or uncalibrated stereo camera systems. \cite{Aghili-Kuryllo-Okouneva-English-2010a,Aghili-2010f,Aghili-Parsa-2009,Aghili-Kuryllo-Okuneva-McTavish-2009,Aghili-2008c,Aghili-Parsa-Martin-2008a}. It followed by development of adaptive visual servoing methods   to deal with various system uncertainties associated with visual servoing law \cite{Xie-Low-He-2017}. Methods for robust vision-based control of robots given limitations of the camera field of view for visual servoing are presented in \cite{Wang-Lang-Silva-2010}.  Image-based visual servoing based on predictive control method has been proposed in \cite{Mcfadyen-Corke-2014} for robust control and guidance of small unmanned aerial vehicles.   Image-based visual servoing using an optimized trajectory planning technique is presented in \cite{Keshmiri-Xie-2017}.   Robotic path planning to achieve robust visual servoing for a specific industrial task is discussed in  \cite{Chen-Wang-Zhao-2018}. The first design concept of a visual servoing system for space robots was presented in \cite{Inaba-Oda-Hayashi-2003} along with experimental results using the Japanese ETS-VII testbed. The iterative recursive least-square or Extended Kalamn filter methods have been proposed to
perform motion estimation of a floating object from  the visual information of a stereo-vision system or a laser camera system~\cite{Aghili-Parsa-2007b,Linchter-Dubowsky-2004,Aghili-Parsa-2009}.  Motion planning and control strategies for capture  a non-cooperative target satellite and then stabilized it have been  presented in~\cite{Luo-Sakawa-1990,Stengel-1993,Aghili-2008c,Aghili-2011k,Aghili-2016c,Aghili-2022}.

This paper presents a  visual servoing system  to provide  fault-tolerance and self-tuning capabilities  based on seamless integration of a vision registration algorithm, adaptive stochastic estimator, fault detection and recovery strategies,  as well as optimal motion planning and predictive control \cite{Aghili-2022}. To improve the performance and robustness of the image registration algorithm, a dynamic estimator capturing  evolution of the  relative motion is integrated with the registration algorithm in a closed-loop configuration to achieve recovery after temporary vision failure.  A novel dynamics model in terms of {\em minimum set} inertia parameters is developed to design the {\em constrained estimator} based on derivation of elegant equality constraints that lead to improved  the  degree of the observability of the system.
The metric fit error of the image registration, i.e., the distance between the surface model of the target and point cloud acquired by the vision sensor is used by a fault-detection logic for health monitoring  of the visual feedback. The fault detection and recovery  mechanism in conjunction with adaptively readjusting the observation covariance matrix  allow the overall motion  estimator tunes itself in the face of inaccurate or erroneous vision data. It also makes continuous pose estimation  possible even in the presence of temporary sensor failure or short-term image occlusion. The estimator also provides accurate knowledge of the state variables and inertial parameters to an optimal guidance and control system for rendezvous prediction  and  motion planning purposes so that the motion of the end-effector can be carefully planned and executed given multiple physical and environmental constraints. This work encompasses significant improvement in the performance and robustness of visual servoing system  through a series of new technical contributions as follows. Firstly, a novel dynamics model in terms of {\em minimum set} inertia parameters is developed and subsequently a {\em constrained estimator} is designed to elevate the degree of the observability of the system. A rule-based logic system has been incorporated in the motion estimation/prediction process  to  improve the robustness of the automated rendezvous and capture. The second technical contribution of this paper is the development of exact solution of the optimal robot  trajectory for capturing moving objects using {\em hard constraint method} instead of {\em soft constraint method} based on penalty function used in \cite{Aghili-2011k}. Notice that the relating methods for interception trajectories  based on penalty function method require introducing an augmented performance index and thus changing the original cost function of the optimal control problem.

\section{Fault-Tolerant Motion Estimation Based on Vision Data} \label{sec:ICP}
This section describes  the hierarchical estimation and control approach for the fault-tolerant vision-guided robotic systems to capture moving objects. The integrated image registration algorithm and constrained noise-adaptive Kalman filter provides consistent estimation of the states and parameters of the target from the stream of visual data \cite{Aghili-Salerno-2016}. The health of the visual feedback is assessed based on the metric fit error of the image registration in order to make decision whether or not the visual data has to be incorporated  in the estimation process.  Transition on the convergence of parameter estimation is monitored from the magnitude of the estimator covariance matrix while the robot remains stand-still.  The parameters are passed to the optimal guidance consisting of the optimal rendezvous and trajectory planning modules at the time of initial convergence $t_1$. The time of interception of the capturing robot and the target is determined by the optimal rendezvous module and subsequently the optimal trajectories given  physical and operational constraints are generated by the optimal planner.

The iterative closest point (ICP) algorithm is  the most popular algorithm for solving registration process of 3D CAD model and point cloud generated by a 3D vision sensor. Suppose that we are given with a set
of acquired 3D points data  ${\cal C}_k=\{\bm c_{1_k} \cdots \bm c_{m_k} \}$ at discrete time $t_k$ that corresponds to a single shape represented by model set ${\cal M}$. Here, vector $\bm c_{i_k} \in\mathbb{R}^3$  represents the coordinate of $i$th single point from the point cloud at epoch $k$. For each point $\bm c_{i_k}$, there exists at least one point on
the surface model ${\cal M}$, say $\bm d_{i_k}$, which is closer to $\bm c_{i_k}$ than other points. Therefore, one should be able to populate the data set ${\cal D}_k = \{\bm d_{1_k} \cdots \bm d_{m_k} \}$ representing  all corresponding points from an optimization process ~\cite{Simon-Herbert-Kanade-1994}. Then, the fine alignment $\{ \bar{\bm\rho}_k, \; \bar{\bm\eta}_k \}$, where $\bar{\bm\rho}_k$ represents the distance and  unit-quaternion $\bar{\bm\eta}_k$ represents the orientation, can be resolved  to minimize the distance between the data sets
 ${\cal C}_k$ and ${\cal D}_k$ through the the following least squares programming~\cite{Besl-Mckay-1992}
\begin{equation} \label{eq:min_distance}
\varepsilon_k = \mbox{arg} \min_{\bar{\bm\eta}_k, \bar{\bm\rho}_k} \frac{1}{m} \sum_{i=1}^m \| \bm A(\bar{\bm\eta}_k) \bm c_{i_k} +
\bar{\bm\rho}_k - \bm d_{i_k} \|^2.
\end{equation}
Here, variable $\varepsilon_k$ is called ICP metric fit error, and  $\bm A(\bm\eta)$ is the rotation matrix corresponding to quaternion $\bm\eta$
\begin{equation} \label{eq:R}
\bm A(\bm\eta) =\bm I + 2 \eta_o [\bm\eta_v \times] + 2 [\bm\eta_v \times]^2,
\end{equation}
where $\bm\eta_v$ and $\eta_o$ are the vector and scaler parts of the
quaternion, i.e., $\bm\eta=[\bm\eta_v^T \; \eta_o]^T$, $[\cdot \times]$ denotes the matrix form of the cross-product,
and $\bm I$ denotes  identity matrix with adequate dimension. The complete cycle of 3D registration process with incorporation of predicted pose will be later discussed in Section~\ref{sec:3D-registration}. At this point, it suffices to assume instantaneous pose at epoch $k$ as a function of the point cloud set ${\cal C}_k$, i.e.,
\begin{equation} \label{eq:ICP_outcome}
\begin{array}{l}\mbox{outcome of ICP} \\ \mbox{cycle at $ t_k $} \end{array}:= \left\{\begin{array}{c} \bar{\bm\rho}_k({\cal C}_k, \bm\rho_k^{(0)}, \bm\eta_k^{(0)}) \\ \bar{\bm\eta}_k({\cal C}_k, \bm\rho_k^{(0)}, \bm\eta_k^{(0)}) \\  \varepsilon_k \end{array} \right\}.
\end{equation}
Here,  $\bm\rho_k^{(0)}$ together with $\bm\eta_k^{(0)}$ represent initial guess for the coarse alignment to be rendered by a prediction estimation process. The remainder of this section describes development of a fault-tolerant dynamics estimator based on \eqref{eq:min_distance} and \eqref{eq:ICP_outcome}.

The definition of the coordinate frames used for the system of the vision guided manipulator and a moving target is given as follows. The camera coordinate frame  is $\{{\cal A } \}$, while frames $\{{\cal B} \}$ and $\{{\cal C}\}$ are both attached to the body of the target and their origins coincide respectively with the center-of-mass (CoM) and  the
grapple fixture located at distance $\bm\varrho$ from the CoM. We also assume that frame $\{ B \}$ is aligned with principal axes of the body. The vision system gives a noisy measurement of the position and orientation of coordinate $\{{\cal C}\}$  with respect to the coordinate
frame $\{{\cal A}\}$ represented by variables $\bm\rho$ and unit quaternion $\bm\eta$. Suppose unit quaternions $\bm\mu$ and $\bm q$, respectively, represent the misalignment between coordinates $\{{\cal B}\}$ and $\{{\cal C}\}$ and the orientation
of coordinates frame $\{{\cal B}\}$ respect to $\{{\cal A}\}$.
Then, $\bm\eta$ can be considered as two
successive orientations as
\begin{equation} \label{eq:muOtimesq}
\bm\eta = \bm\mu \otimes \bm q, \quad \mbox{where} \quad  \bm\mu \otimes = \mu_o \bm I + \bm\Omega(\bm\mu_v)
\end{equation}
is the quaternion product operator, and
\begin{equation} \label{eq:Omega}
 \bm\Omega(\bm\mu_v)=\begin{bmatrix} -[\bm\mu_v
\times] & \bm\mu_v \\ - \bm\mu_v^T & 0 \end{bmatrix}.
\end{equation}
Since the target rotates with
angular velocity $\bm\omega$, quaternion $\bm\mu$ is a constant whereas $\bm\eta$ and $\bm q$ are time-varying variables. Now let us define the following non-dimensional inertia parameters
\begin{equation} \label{eq:p}
\sigma_1 = \frac{I_{yy}- I_{zz}}{I_{xx}}, \quad
\sigma_2 = \frac{I_{zz}- I_{xx}}{I_{yy}}, \quad
\sigma_3 = \frac{I_{xx}- I_{yy}}{I_{zz}},
\end{equation}
where $I_{xx}$, $I_{yy}$, and $I_{zz}$ denote the target's principal moments
of inertia. The principal moments of inertia take only  positive values and and they must also satisfy the triangle inequalities. Thus, the following inequalities are in order
\begin{align} \notag
& I_{xx}, I_{yy}, I_{zz} >0 \\ \notag
& I_{xx} + I_{yy} > I_{zz}, \quad I_{yy} + I_{zz} > I_{xx}, \quad I_{zz} + I_{xx} > I_{yy}.
\end{align}
The above inequalities imply the dimensionless inertia parameters must be bounded by
\begin{equation} \label{eq:sigma_bounds}
-1 < \sigma_1, \sigma_2, \sigma_3 < 1
\end{equation}
Also by inspection one can show that the following nonlinear constraint between the dimensionless parameters is in order
\begin{equation} \label{eq:sigma_constraint}
\Gamma(\bm\sigma) = \sigma_1 + \sigma_2 + \sigma_3 + \sigma_1\sigma_2\sigma_3  =0,
\end{equation}
which means that the dimensionless parameters are not independent. Consider the set of independent dimensionless inertia parameters $\bm\sigma = [\sigma_1 \; \;  \sigma_2]^T$. Then, from \eqref{eq:sigma_constraint} one can readily derive the third variable by
\begin{equation}
\sigma_3= - \frac{\sigma_1 + \sigma_2}{1 + \sigma_1 \sigma_2}
\end{equation}
Inequalities \eqref{eq:sigma_bounds} can be concisely described by the following vector inequality 
\begin{equation} \label{eq:Dsigma<1}
\bm D \bm\sigma < \bm 1, \quad \mbox{where} \quad  \bm D =\begin{bmatrix} \bm I \\  -\bm I \end{bmatrix}
\end{equation}
where $\bm 1=[1\; 1\; 1 \; 1]^T$ is the vector of one . Now, we are ready to transcribe the Euler's rotation equations  in terms of the independent dimensionless parameters $\bm\sigma$, i.e.,
\begin{subequations}
\begin{equation} \label{eq:dot_omega}
\dot{\bm\omega} = \bm\phi(\bm\omega, \bm\sigma) + \bm B(\bm\sigma)
\bm\epsilon_{\tau}
\end{equation}
where $\bm\epsilon_{\tau} = \bm\tau/{\rm tr}(\bm I_c)$ is the angular acceleration disturbance, $\bm I_c =\mbox{diag}(I_{xx}, I_{yy},I_{zz})$ is the inertia tensor,  ${\rm tr}(\cdot)$ is the trace operator,
\begin{align} \notag
\bm B(\bm \sigma) & = \bm I + \begin{bmatrix}\frac{2+\sigma_1 \sigma_2 + \sigma_1}{1 - \sigma_2} & 0 & 0 \\
0 & \frac{2+\sigma_1 \sigma_2 - \sigma_2}{1 + \sigma_1} & 0 \\
0 & 0 & \frac{2+\sigma_1 -\sigma_2 }{1 + \sigma_1 \sigma_2} \end{bmatrix}, \\
\bm\phi(\bm\omega, \bm\sigma) &= \begin{bmatrix} \sigma_1 \omega_y \omega_z \\ \sigma_2 \omega_x \omega_z \\
-\frac{\sigma_1 + \sigma_2}{1 + \sigma_1 \sigma_2} \omega_x \omega_y  \end{bmatrix}.
\end{align}
\end{subequations}
Consider the following state vector pertaining to both states  and the associated dynamic parameters
\begin{equation} \label{eq:x}
\bm x=[\bm q_v^T \; \bm\omega^T \; \bm\rho_o^T \; \dot{\bm\rho}_o^T \; \underbrace{\bm\sigma^T \; \bm\varrho^T \;  \bm\mu_v^T }_{\rm parameters} \; ]^T.
\end{equation}
Then, from the well-known relationship between the time derivative of quaternion and angular rate, equations of linear and angular accelerations, and assuming constant parameters, one can describe the system dynamic in the following compact form
\begin{equation}
\dot{\bm x} = \bm f(\bm x, \bm\epsilon), \qquad \bm f(\bm x, \bm\epsilon) = \begin{bmatrix} \frac{1}{2} \mbox{vec} \big( \bm\Omega(\bm\omega) \bm q \big) \\ \bm\phi(\bm\omega, \bm\sigma) +\bm B(\bm\sigma) \bm\epsilon_{\tau} \\
\dot{\bm\rho}_o \\
\bm\epsilon_f  \\ \bm 0 \end{bmatrix},
\end{equation}
where function $\mbox{vec}(\cdot)$ returns the vector part of quaternion, $\bm\epsilon_f$ is the process noise due to force disturbance, and vector $\bm\epsilon= [\bm\epsilon_{\tau}^T \;\; \bm\epsilon_f^T]^T$ represent the overall process noise.
 Suppose $\hat{\bm q}$ represents the estimated quaternion and subsequently define small quaternion variable $\delta \bm q = \bm q \otimes \hat{\bm q}^{-1}$ to be  used as the states of linearized system, where $\hat{\bm q}^{-1}$ is quaternion inverse, i.e., $\hat{\bm q} \otimes \hat{\bm q}^{-1}=[0\; 0 \; 0 \; 1]^T$. Then, the linearized process dynamics can be described by
\begin{subequations} \label{eq:dot_deltax}
\begin{equation}
\delta \dot{\bm x} = \bm F \delta \bm x + \bm G \bm\epsilon,
\end{equation}
\begin{align} \label{Eq:F}
\bm F &=
\begin{bmatrix} - [\hat{\bm\omega} \times] & \frac{1}{2} \bm I & \bm 0 & \bm 0 & \bm 0 & \bm 0 & \bm 0 \\
\bm 0& \left(\frac{\partial\bm\phi}{\partial\bm\omega}\right)_{\!\! \hat{\bm x}}   & \bm 0 & \bm 0 & \left(\frac{\partial\bm\phi}{\partial\bm\sigma}\right)_{\!\! \hat{\bm x}}  & \bm 0 & \bm 0 \\
\bm 0 & \bm 0& \bm 0 & \bm I & \bm 0 & \bm 0 & \bm 0   \\
\bm 0 & \bm 0 & \bm 0 & \bm 0 & \bm 0 & \bm 0 & \bm 0
\end{bmatrix} \\ \label{Eq:G}
\bm G &= \begin{bmatrix} \bm 0 & \bm 0 \\
\bm B(\hat{\bm\sigma}) & \bm 0 \\
\bm 0& \bm 0\\
\bm 0& \bm I \\
\bm 0 & \bm 0
\end{bmatrix} \\
\frac{\partial\bm\phi}{\partial\bm\omega} &=
\begin{bmatrix} 0 & \sigma_1 \omega_z & \sigma_1  \omega_y \\ \sigma_2  \omega_z & 0 &
\sigma_2 \omega_x \\ -\frac{\sigma_1 +\sigma_2 }{1 + \sigma_1 \sigma_2} \omega_y & -\frac{\sigma_1 + \sigma_2}{1 + \sigma_1 \sigma_2} \omega_x & 0
\end{bmatrix} \\
\frac{\partial\bm\phi}{\partial\bm\sigma} & = \begin{bmatrix} \omega_y \omega_z & 0 \\ 0 & \omega_x \omega_z  \\ \frac{\sigma_2^2 -1}{(1+ \sigma_1 \sigma_2)^2}  \omega_x \omega_y & \frac{\sigma_1^2 -1}{(1+ \sigma_1  \sigma_2)^2} \omega_x \omega_y \end{bmatrix}.
\end{align}
\end{subequations}

Defining quaternion variations $\delta \bm\mu=\hat{\bm\mu}^{-1} \otimes \bm\mu$ and $\delta \bm\eta = \delta \bm\mu \otimes \delta \bm q$, one can readily establish the relationship between the measured quaternion and its variation through the following identity $\bm\eta = \hat{\bm\mu} \otimes \delta \bm\eta \otimes \hat{\bm q}$, and hence $\delta \bar{\bm\eta} = \hat{\bm\mu}^{-1} \otimes \bar{\bm\eta} \otimes \hat{\bm q}^{-1}$. Then, by virtue of \eqref{eq:ICP_outcome} and \eqref{eq:muOtimesq}, the observation equations can be written as
\begin{equation} \label{eq:h_nonlin1}
\bm z = \begin{bmatrix} \bar{\bm\rho} \\ \delta \bar{\bm\eta}_v \end{bmatrix}= \bm h(\delta \bm x) + \bm v
\end{equation}
where vector $\bm v$ represents measurement noise, and
\begin{equation} \label{eq:h_nonlin}
\bm h (\delta \bm x) = \begin{bmatrix} \bm\rho_o + \bm A(\delta \bm q \otimes \hat{\bm q}) \bm\varrho \\ \mbox{vec} \big( \delta \bm\mu \otimes \delta \bm q \big) \end{bmatrix}.
\end{equation}
Finally, from the first-order approximation of the relationship $\bm A(\delta \bm q \otimes \hat{\bm q}) \approx 2 \bm A(\hat{\bm q})[\delta \bm q_v \times]$ and $\mbox{vec}(\delta \bm\mu \bm \otimes \delta \bm q) \approx - \delta \bm\mu_v \times \delta \bm q_v + \delta \bm q_v + \delta \bm\mu_v$, one can derive the observation sensitivity
matrix $\bm H= \bm H (\delta \hat{\bm x})$ in the following form
\begin{equation} \notag
\bm H =
\begin{bmatrix} -2\bm A(\hat{\bm q})[\hat{\bm \varrho} \times] &\bm 0
& \bm I & \bm 0  & \bm 0 & \bm A(\hat{\bm q}) &\bm 0\\
\bm I - [\delta \hat{\bm\mu}_v \times ]  &\bm 0  &\bm 0 &\bm 0  & \bm 0& \bm 0 & \bm I + [\delta \hat{\bm q}_v \times ]
\end{bmatrix}.
\end{equation}

\subsection{Adaptive Constrained EKF}
Define $\delta \hat{\bm x}_k^-$  and $\delta \hat{\bm x}_k^+$ as the {\em a prioir} and {\em a posteriori} estimates of the state vector at time $t_k$ and their corresponding estimation errors $\delta \tilde{\bm x}_k^-= \delta \bm x_k - \delta \hat{\bm x}_k^-$ and $\delta \tilde{\bm x}_k^+=\delta \bm x_k - \delta  \hat{\bm x}_k^+$ with associated covariances  $\bm P_k^-=E[\delta \tilde{\bm x}_k^- \delta \tilde{\bm x}_k^{-T}]$ and $\bm P_k^+=E[\delta \tilde{\bm x}_k^+ \delta \tilde{\bm x}_k^{+T}]$, where $E[\cdot]$ is the expected operator. The estimation update is
\begin{equation} \label{eq:innovation}
\delta \hat{\bm x}_k^+ = \delta \hat{\bm x}_k^- + \bm K_k (\bm z_k - \bm h(\delta \hat{\bm x}_k^-))
\end{equation}
The Kalman filter gain  minimizes the performance index $E(\| \delta \tilde{\bm x}_k^+ \|^2)=\mbox{tr}(\bm P_k^+)$ subject to the state constraints \eqref{eq:Dsigma<1}. Therefore, according to the Joseph formula, the constrained Kalman filter is the solution of the following optimization programming
\begin{align*}
\min_{\bm K_k}  \mbox{tr} & \big((\bm I - \bm K_k \bm H_k) \bm P_k^- (\bm I + \bm K_k \bm H_k)^T + \bm K_k \bm R_k \bm K_k^T  \big) \\
\mbox{subject to:} & \quad \bm D \bm\hat{\bm\sigma}_k^+ \prec \bm 1
\end{align*}
The gain projection technique can be applied to impose the inequality constraints for the estimation process \cite{Gupta-Hauser-2007,Teixeira-Chandrasekar-2008}. In this method if the {\em aprioir} estimate  satisfies the constraints but the unconstrained {\em aposteriori} estimate $\delta \hat{\bm x}_k^+$ does not satisfy them, then  the latter can be projected in the direction of the former until it reaches the constraint boundary. This effectively gives modified Kalman gain as follow
\begin{equation}
\bm K_k = \bm\beta_k \bm K^u_k,
\end{equation}
Here, $\bm K^u$ is the standard unconstrained Kalman gain and $\bm\beta_k=\mbox{diag}(1, 1, \cdots, \beta_{1_k}, \beta_{2_k}, \cdots, 1,1)$ where
\begin{equation} \notag
\beta_{i_k} =\left\{ \begin{array}{ll}
\mbox{sgn}(\bm k_{i_k}^T \bm e_k) -\frac{\hat{\sigma}_{i_k}^-}{\bm k_{i_k}^T \bm e} & \quad \mbox{if} \quad |\bm k_{i_k}^T \bm e_k| >1 \\
1 & \quad \mbox{otherwise} \end{array} \right.  \quad i=1,2
\end{equation}
with $\bm k^T_{1_k}$ and $\bm k^T_{2_k}$ being the last two row vectors of the unconstrained gain matrix. The unconstrained Kalman gain is given by the standard formula
\begin{subequations}
\begin{align}
\bm K_k^u & = {\bm P}_k^- \bm H_k^T {\bm S}_k^{-1} \\ \label{eq:Sk}
\bm S_k &= \bm H_k \bm P_k^- \bm H_k^T + \bm R_k .
\end{align}
Subsequently the state covariance matrix is updated from
\begin{equation}
{\bm P}_k^+ = \big( \bm I - \bm K_k \bm H_k  \big) {\bm P}_k^-
\end{equation}
\end{subequations}
followed by the propagation of the state and covariance matrix
\begin{subequations} \label{eq:KF_propagate}
\begin{align}\label{eq:state-prop}
\hat{\bm x}_{k+1}^- & = \hat{\bm x}_{k}^+ +
\int_{t_k}^{t_{k}+t_{\Delta}} \bm f(\bm x, \bm
0)\,{\text d} \tau\\ \label{eq:cov-prop}
{\bm P}_{k+1}^-&= \bm\Phi_{k} \bm P_{k}^+ {\bm\Phi}_{k}^T + \bm Q_{k}
\end{align}
\end{subequations}
The covariance matrices $\bm Q_{k}$ and $\bm R_{k}$ in expressions \eqref{eq:cov-prop} and \eqref{eq:Sk} are  associated with process noise and vision sensor noise. A priori knowledge of the  measurement noise covariance is not usually available because of unexpected noise and disturbance associated with 3D vision systems (such as  laser range sensors). This covariance matrix  can be readjusted in real-time from averaging the sequence of either the innovation matrix or the residual matrix \cite{Mehra-1970,Wang-2000,Gao-Wei-Zhong-Subic-2015,Aghili-Salerno-2016}. Consider the residual sequence
\begin{equation} \notag
{\bm e}_k = \bm z_k - \bm H_k \delta \hat{\bm x}_k^+,
\end{equation}
which is the difference between pose measurement and the pose calculated from  {\em a priori} state. Subsequently, the associated  {\em innovation covariance matrix} can be estimated from averaging inside the moving window of size $w$, which can be empirically selected, i.e.,
\begin{equation} \label{eq:S_batch2}
{\bm\Sigma}_k  = \frac{1}{w}\sum_{i=1}^{w} {\bm e}_{k-i}
{\bm e}_{k-i}^{T}
\end{equation}
Then, the estimated observation covariance  is related to innovation covariance matrix \cite{Wang-2000} by
\begin{equation} \label{eq:R_resudua2}
\hat{\bm R_k} \approx {\bm\Sigma}_k + \bm H_k {\bm P}_k^+ \bm H_k^T
\end{equation}
Alternatively, the innovation covariance matrix \eqref{eq:S_batch2} can be recursively updated by
\begin{equation} \label{eq:S_recursive}
{\bm\Sigma}_{k+1} = \left\{ \begin{array}{ll} \frac{k-1}{k} {\bm\Sigma}_{k} + \frac{1}{k} \bm e_{k}  \bm e_{k}^T & \quad \text{if} \quad k<w\\
{\bm\Sigma}_{k} + \frac{1}{w} \Big( \bm e_{k} \bm e_{k}^T -
\bm e_{k-w}\bm e_{k-w}^T \Big) & \quad \text{otherwise}
\end{array} \right.
\end{equation}

\subsection{Observability Analysis}

Only when a system is observable, then its states can be uniquely determined from the measurements regardless of the initial conditions. However, if a system is not fully observable, the  quality of state estimation may be adversely affected by the initial values or the the stability and convergence of the estimator may not be ascertained. A time-varying system is considered to be fully observable if its Observability Gramian Matrix (OGM) is not ill-conditioned \cite{Krener-Ide-2009,Yu-Cui-Zhu-2014,Aghili-2010s}. Thus the degree of the observability of a system can be assessed by examining the condition number of its  OGM. In other words, the condition number of OGM reveals the degree of the observability  of the EKF estimator using the modeling technique. The discrete Observability Gramian matrix can be computed by:
\begin{equation}
\bm W_{O_k} = \sum_{j=1}^k \bm\Phi_{j/0}^T \bm H_j^T \bm H_j \bm\Phi_{j/0},
\end{equation}
where $\bm\Phi_{j/0}= \prod_{i=2}^j \bm\Phi_{i/i-1}$.  The above equation can be also equivalently written in the following recursive form
\begin{align} \notag
\bm W_{O_k} &= \bm W_{O_{k-1}} + \bm\Phi_{k/0}^T \bm H_k^T \bm H_k \bm\Phi_{k/0} \\
 \bm\Phi_{k/0} &= \bm\Phi_{k} \bm\Phi_{k-1/0}
\end{align}
The rank of ill-conditioning of the OG matrix can be quantitatively  represented by its condition number, i.e.,  a very large condition number means the matrix is ill-conditioned and thus not full-rank. 

\section{Fault-Tolerant Vision Data Registration Using Motion Prediction}\label{sec:3D-registration}

The ICP registration is an iterative process that estimates the pose in two steps: i) Find the corresponding points on the model set assuming a coarse-alignment pose is given; ii) Resolve a fine-alignment pose corresponding to the two data sets by minimizing the sum of squared distances. Suppose $\{\bm q^{(n)}, \bm\rho^{(n)} \}$ represent rigid transformation at $n$th cycle of the ICP iteration and that the initial transformation $\{\bm q_k^{(0)}, \bm\rho_k^{(0)} \}$ is obtained from prediction of the states at the propagation step by:
\begin{equation} \label{eq:initial_pose}
\left\{ \begin{array}{l}
{\bm\eta}_k^{(0)} = \hat{\bm\mu}_k^- \otimes \hat{\bm q}_k^- \\
{\bm\rho}_k^{(0)} = \hat{\bm\rho}_{o_k}^- + \bm A(\hat{\bm q}_k^-) \hat{\bm\varrho}_k
\end{array} \right.
\end{equation}
Then, the problem of finding the correspondence between the two sets at the $n$th iteration can be formally expressed by
\begin{equation} \label{eq:ci}
\bm d_{i_k}^{(n)} = \mbox{arg} \min_{\bm d_j \in {\cal M} } \|\bm A \big({\bm\eta}_k^{(n)} \big) \bm c_{i_k}
+ {\bm\rho}_k^{(n)} - \bm d_j^{(n)} \| \quad \forall i=1,\cdots,m,
\end{equation}
and subsequently set ${\cal D}_k^{(n)}$ is formed. Now, we have two independent sets of 3D points ${\cal C}_k$  and ${\cal
D}_k^{(n)}$ both of which corresponds to the same shape but they may not completely coincide due to a rigid-body transformation. The next problem involves finding  the transformation represented by fine alignment $\{ {\bm\rho}_k^{(n+1)}, \; {\bm\eta}_k^{(n+1)} \}$ which
minimizes the distance between these two data sets~\cite{Besl-Mckay-1992}. That is  $ \forall \bm d_{i_k}^{(n)} \in {\cal D}_k^{(n)}, \bm c_{i_k}
\in {\cal C}_k$, we have
\begin{align} \label{eq:ICP}
& \varepsilon_k^{(n)} =  \min_{{\bm\eta}_k^{(n+1)}, {\bm\rho}_k^{(n+1)}} \frac{1}{m} \sum_{i=1}^m \| \bm A({\bm\eta}_k^{(n+1)}) \bm c_{i_k} +
{\bm\rho}_k^{(n+1)} - \bm d_{i_k}^{(n)} \|^2 \\ \notag & \qquad \mbox{subject to} \quad  \| \bar{\bm\eta}_k^{(n+1)} \| =1
\end{align}
where residual $\varepsilon_k^{(n)}$ is the metric fit error at the $n$-th iteration \cite{Horn-1987,Besl-Mckay-1992}.
To solve the above least-squares
minimization problem, we define the
cross-covariance matrix of the sets ${\cal C}_k$ and ${\cal D}_k^{(n)}$
by
\begin{equation}
\bm N^{(n)} = \mbox{cov}({\cal C}_k, {\cal D}_k^{(n)}) = \frac{1}{m} \sum_{i} \bm
c_{i_k} \big(\bm d_{i_k}^{(n)}\big)^T - {\bm c}_{o_k} \big({\bm d}_{o_k}^{(n)} \big)^T,
\end{equation}
where ${\bm c}_{o_k} = \frac{1}{m} \sum_{i} \bm c_{i_k}$ and ${\bm d}_{o_k}^{(n)}
= \frac{1}{m} \sum_{i} \bm d_{i_k}^{(n)}$ are the corresponding centroids of the point cloud data set. Let us also define the following symmetric
matrix constructed from  $\bm N$
\begin{equation} \notag
\bm M^{(n)} = \begin{bmatrix}
\mbox{tr}(\bm N^{(n)}) & \big(\bm n^{(n)} \big)^T \\
\bm n^{(n)} & \bm N^{(n)} + \big(\bm N^{(n)}\big)^T - \mbox{tr}\big(\bm N^{(n)} \big) \bm I
\end{bmatrix},
\end{equation}
where $\bm n^{(n)}=[N_{23}^{(n)} - N_{32}^{(n)}, N_{31}^{(n)} -N_{13}^{(n)}, N_{12}^{(n)}-N_{21}^{(n)}]^T$. Then, it
has shown in  \cite{Horn-1987} that minimization problem \eqref{eq:ICP} can be equivalently transcribed by the following quadratic programming
\begin{equation} \label{eq:quadratic}
\max_{\| {\bm\eta}_k^{(n+1)} \| =1} \big({\bm\eta}_k^{(n+1)} \big)^T \bm M^{(n)} \big( {\bm\eta}_k^{(n+1)} \big),
\end{equation}
which has the following closed-form
\begin{equation} \label{eq:q_rime}
\begin{array}{cc}
{\bm\eta}_k^{(n+1)} &={\mbox{eigenvector} \big(\bm M^{(n)} \big)}\\
& \lambda_{\rm max} \big(\bm M^{(n)} \big)
\end{array}
\end{equation}
Next, we can proceed with computation of the translation by
\begin{equation}  \label{eq:r_rime}
{\bm\rho}_k^{(n+1)} = {\bm d}_{o_k}^{(n)} - \bm A({\bm\eta}_k^{(n+1)}) {\bm c}_{o_k}
\end{equation}
The fine alignment obtained from  \eqref{eq:q_rime} and \eqref{eq:r_rime} can be used in Step I and then  continue the iterations until the residual error $\varepsilon_n^{(n)}$ becomes less
than the pre-specified  threshold $\varepsilon_{\rm th}$. That is
\begin{equation} \notag
\left\{ \begin{array}{ll}
\mbox{if} \quad \varepsilon_k^{(n)} < \varepsilon_{\rm th}   \quad & \bar{\bm\eta}_k = \bm\eta_k^{(n+1)}, \quad \bar{\bm\rho}_k = \bm\rho_k^{(n+1)}, \\
& \mbox{and exit the iteration}\\
\mbox{otherwise} & \mbox{go to \eqref{eq:ci} and use $\bm\eta^{(n+1)}$ and $\bm\rho_k^{(n+1)}$} \\
&  \mbox{as the new coarse alignment}
\end{array} \right.
\end{equation}
Suppose $n_{\rm max}$  represent the maximum number of iterations specified by a user. Then, the ICP loop is considered convergent if the matching error $\varepsilon_k^{(n)}$ is less than $\varepsilon_{\rm th}$  and $n \leq n_{\rm max}$. Otherwise, the ICP iteration at epoch $k$ is considered a failure. We should incorporate the refinement  rigid-body transformation  in the KF innovation sequence only when the ICP becomes convergent. To this goal, we introduce the following flagged variable to indicate wether the registration process at epoch $k$  is healthy or faulty
\begin{equation}
\gamma_k = \left\{ \begin{array}{ll} 1  \quad & \mbox{if} \quad  \varepsilon_k^{(n)} < \varepsilon_{\rm th} \quad \wedge \quad n \leq n_{\rm max} \\
0 & \mbox{otherwise} \end{array} \right.
\end{equation}
It is worth noting that although the necessary condition for convergent of ICP to a correct solution is small metric fit  error, the validity of the registration algorithm can be also examined by comparing the pose resolved by matching the date sets with the predicted pose obtained from  the dynamics model, i.e., $\bm\alpha_k =\bm z_k - \bm h(\delta \hat{\bm x}_k^-)$ calculated in  the innovation step \eqref{eq:innovation}. Then, the condition for detecting ICP fault ($\gamma_k=0$) can be described by
$\varepsilon \geq \varepsilon_{\rm th}$ and $\| \bm\alpha_k \|_W \geq \alpha_{\rm th}$, where $\alpha_{\rm th}$  is the pose error threshold, and $\| \cdot \|_W$ denotes weighted Euclidean norm. Here, the weight matrix can be properly selected as $\bm W= \mbox{diag}(\bm I , \; L \bm I  )$, where $L$ is the called the characteristic length which is empirically obtained as the ratio of the maximum linear and angular errors \cite{Aghili-2010p,Aghili-Parsa-2008b}. Once ICP fault is detected, the pose-tracking fault recovery is rather straightforward. To this end, the Kalman filter gain is set to
\begin{equation} \label{eq:K_phi}
\bm K_k  = \gamma_k \bm\beta_k {\bm P}_k^- \bm H_k^T \hat{\bm S}_k^{-1}
\end{equation}
Clearly $\bm K_k = \bm 0$ when ICP fault is detected, in which case the observation information is not incorporated in the estimation process  for updating the state and the estimator covariance, i.e., $\gamma_k=0 \; \Longrightarrow \; \bm K_k = 0$ and thus $\hat{\bm x}_k^+ = {\bm x}_k^-$ and $\bm P_k^+ = \bm P_k^-$. In other words, the estimator relies on the dynamics model for pose estimation until ICP becomes convergent again.

\section{Optimal Rendezvous \& Guidance for Visual Servoing} \label{sec:guidance}
This section presents an optimal robot guidance approach for rendezvous and smooth interception of a moving object based on visual feedback. The position of the end-effector and the grasping point are denoted by  $\bm r$ and $\bm\rho$, respectively.
The end-effector and the grasping point are expected to arrive at a rendezvous-point
simultaneously with the same velocity in order to avoid impact at
the time of grasping. Suppose the optimal trajectory is manifested  by the
following second order system
\begin{equation} \label{eq:sys_xr}
\ddot{\bm r} =\bm u,
\end{equation}
which can be formally rewritten as $\dot{\bm\xi} =[ \dot{\bm r}^T  \; \bm u^T ]^T$ with state vector $\bm\xi^T=[\bm r^T \; \dot{\bm r}^T]$, and terminal condition $\bm r(t_f) =\bm\rho(t_f)$ and $\dot{\bm r}(t_f) =\dot{\bm\rho}(t_f)$ at terminal time $t_f$. In the following analysis, we seek a time-optimal solution to the input $\bm u$ subject to the acceleration limit $\|\ddot{\bm r} \|\leq a_{\rm max}$ and the aforementioned terminal constraints, i.e.,
\begin{subequations}
\begin{align}
\mbox{minimize}  &  \qquad\int_t^{t_f} 1 \; d \tau\\ \label{eq:a_max}
\mbox{subject to:}   & \qquad \| \bm u(\tau) \| \leq a_{\rm max}  \qquad t \leq \tau \leq t_f \\ \label{eq:terminal}
& \qquad \bm r(t_f) = \bm\rho(t_f), \qquad \dot{\bm r}(t_f) = \dot{\bm\rho}(t_f)
\end{align}
\end{subequations}
Defining the vector of Lagrangian multiplier as $\bm\lambda$, one can write the expression of the system Hamiltonian as
\[ H = 1 + \bm\lambda^T \dot{\bm\xi} \]
According to the optimal control
theory~\cite{Anderson-Moore-1990}, the costate  must satisfy
\begin{equation} \notag
\dot{\bm\lambda} = -\frac{\partial H}{\partial \bm\xi} \quad \mbox{hence} \quad \bm\lambda^* = \begin{bmatrix} \bm c_1 \\ - \bm c_1 \tau + \bm c_2 \end{bmatrix} \quad \forall \tau\in[t, \; t_f],
\end{equation}
where $^*$ indicates optimal values, vectors $\bm c_1$ and $\bm c_2$ are unknown constant vectors to be found from the boundary conditions \eqref{eq:terminal}. Thus
\begin{equation} \label{eq:H*}
H(\bm\xi^*, \bm\lambda^*, \bm u) = 1 + \bm c_1^T \dot{\bm r} + (-\bm c_1 \tau + \bm c_2 )^T \bm u.
\end{equation}
The Pontryagin's principle dictates that the optimal input $\bm u^*$ satisfies
\begin{equation} \notag
\bm u^* = \mbox{arg} \min_{\bm u} H(\bm\xi^*, \bm\lambda^*, \bm u).
\end{equation}
Defining vector $\bm p = -\bm c_1 \tau + \bm c_2 $, one can infer from \eqref{eq:H*} that the optimal control input $\bm u^*$ minimizing the Hamiltonian should be aligned with unit vector $-\bm p/\| \bm p \|$. Therefore, in view of the acceleration limit constraint \eqref{eq:a_max}, the optimal input must take the following structure
\begin{equation} \label{eq:u}
\bm u^*  =-\frac{-\bm c_1 \tau  + \bm c_2}{\| -\bm c_1 \tau  + \bm c_2 \|} a_{\rm max} \quad \forall \tau\in[t, \; t_f]
\end{equation}
However, the optimal terminal time $t_f$ along with vectors $\bm c_1$ and $\bm c_2$ remain to be found. The
{\em optimal Hamiltonian} calculated at optimal point $\bm u^*$ and $\bm\xi^*$
must also satisfy
\[ \frac{d}{d \tau} H(\tau) =0 \]
and hence
\begin{equation} \notag
1+ \bm c_1^T \dot{\bm r} + \|  \bm c_1 \tau - \bm c_2 \| a_{\rm max} = \mbox{const}.
\end{equation}
Applying the initial and final conditions  to the above equation yields
\begin{align} \notag
\Delta  H(\bm c_1, \bm c_2, t_f) &=  \big(\| \bm c_1 t - \bm c_2 \| -  \| \bm c_1 t_f - \bm c_2 \| \big) a_{\rm max} \\ \label{eq:H=0} & +  \bm c_1^T \big(\dot{\bm\rho}(t_f) - \dot{\bm r}(t_f) \big) =0.
\end{align}
Finally applying the terminal conditions \eqref{eq:terminal} to \eqref{eq:u} and combining the resultant equations with \eqref{eq:H=0}, we arrive at the following error equation in terms of seven unknowns $\bm \chi^T = [\bm c_1^T \;  \bm c_2^T \; t_f ]$.
\begin{equation} \label{eq:phi}
e(\bm \chi) = \left\| \begin{bmatrix} \int_{t}^{t_f} \bm u(\bm c_1, \bm c_2, \nu) d \nu - \dot{\bm\rho}(t_f) \\ \int_{t}^{t_f} \int_{t}^{\mu} \bm u(\bm c_1, \bm c_2, \nu) d \nu d \mu  - \bm\rho(t_f) \\ \Delta  H(\bm c_1, \bm c_2, t_f) \end{bmatrix} \right\| = 0,
\end{equation}
where $\bm\rho(t_f)$ and $\dot{\bm\rho}(t_f)$ are the position and velocity along the target trajectories at the interception, which can be calculated from
\begin{align} \notag
\dot{\bm\rho}(t_f) &=   \int_t^{t_f}  \bm A({\bm q})[ {\bm\omega} \times( {\bm\omega} \times \hat{\bm\varrho}) + \bm\phi({\bm\omega}, \hat{\bm\sigma})\times \hat{\bm\varrho} \; ] d \tau \\
\bm\rho(t_f) &= \int_t^{t_f} \dot{\bm\rho}(\tau) d \tau.
\end{align}
The set of seven nonlinear equations in \eqref{eq:phi} can be solved for seven unknowns $\{\bm c_1, \bm c_2, t_f \}$ by a numerical method. Then the solution can be substituted
into \eqref{eq:u} to resolve the optimal robot trajectories by integration. Clearly, the error $e$ computed at correct variable should vanishes at the terminal time. Therefore, one may call an unconstrained optimization function  (\verb"fminunc" function in Matlab) to find the solution based on a quasi-Newton method \cite{Boyarko-Yakimenko-2011}.

\section{Conclusions}

An adaptive and fault-tolerant  vision-guided robotic system has been presented that allowed precise and smooth capturing of a moving object even if  pose tracking of the vision system was temporary lost. The hierarchical visual servoing was developed by interwoven  integration of a variant of ICP registration, a constrained noise-adaptive Kalman filter, a fault detection logic, and an optimal guidance and control. We proved existence of  elegant equality and inequality constraints associated with dimensionless inertial pertaining  to free-floating object and subsequently developed an innovative dynamics model in terms of a minimum set parameters. Subsequently an adaptive constrained KF estimator was developed for consistent parameter and state estimations by imposing the triangle  inequality while adjusting the observation covariance matrix in real-time. Strategies for fault detection and pose tracking recovery have been implemented based on the prediction and the ICP metric fit errors in conjunction with closed-loop integration of the image registration process and the dynamic estimator. Finally, an optimal rendezvous and guidance scheme was developed which incorporates the processed visual feedback information and then executes optimal robot trajectories  to intercept the grasping point on the moving target with zero relative velocity as quickly as possible subject to  norm limit  of the robot acceleration. 

\bibliographystyle{IEEEtran}

\end{document}